\documentclass[12pt]{iopart}
\usepackage{graphicx}
\expandafter\let\csname equation*\endcsname\relax
\expandafter\let\csname endequation*\endcsname\relax
\usepackage{amsmath}
\usepackage{physics}
\usepackage{txfonts, comment}
\usepackage{algorithm}
\usepackage{algorithmic}
\usepackage{cite}
\usepackage{color}
\usepackage[OT1]{fontenc}
\usepackage{soul}
\newcommand{\ten}{\mathsf{T}}
\newcommand{\rar}{\rightarrow}
\newcommand{\av}[1]{\left \langle {#1} \right \rangle}

\DeclareMathOperator*{\extr}{Extr}

\newcommand{\prior}[1]{p_U^0(#1)}
\newcommand{\priorV}[1]{p_V^0(#1)}
\newcommand{\priorU}[1]{p_U^0(#1)}
\DeclareMathOperator*{\myTr}{Tr}
\begin{document}
\title{Matrix completion based on Gaussian \color{black}parameterized \color{black} belief propagation}
\author{Koki Okajima and Yoshiyuki Kabashima}
\address{Graduate School of Science, The University of Tokyo, Bunkyo, Tokyo 113-0033, Japan  \\} 
\begin{abstract}
We develop a message-passing algorithm for noisy matrix completion problems based on matrix factorization.
The algorithm is derived by approximating message distributions of belief propagation with Gaussian distributions that 
share the same first and second moments.
We also derive a memory-friendly version 
of the proposed algorithm by applying a perturbation treatment 
commonly used in the literature of approximate message passing. 
In addition, a damping technique,
which is demonstrated to be crucial for optimal performance, 
is introduced without computational strain, and the relationship 
to the message-passing version of alternating least squares, 
a method reported to be optimal in certain
settings, is discussed. 
Experiments on synthetic datasets show 
that while the proposed algorithm quantitatively exhibits almost the same 
performance under settings where the earlier algorithm is optimal, 
it is advantageous when the observed datasets 
are corrupted by non-Gaussian noise. 
Experiments on real-world datasets also emphasize the performance 
differences between the two algorithms.
\end{abstract}
\begin{indented}
\item \noindent{\it Keywords}: matrix completion, matrix factorization, belief propagation
\end{indented}
\section{Introduction}
Estimating the elements of a matrix from its sparse, noisy 
observation is a widely studied problem relevant to real-world applications 
such as collaborative filtering and recommender systems. 
Although this problem is ill-defined, manageable results can still be obtained by 
constraining the observed matrix to be of low rank. \\
\indent A primary setting for noisy low-rank matrix completion would be to minimize the 
rank of the matrix subject to constraining the observed and estimated entries by a 
margin $\delta$: 
\begin{equation}\label{eq:rankmin}
      \min_{X\in \mathbb{R}^{N \times M}} {\rm rank}\ X \quad {\rm subject\ to} \quad \abs{x_{ij} - y_{ij}}  \leq \delta , \quad (i,j) \in \Omega,
\end{equation}
where $X = \{x_{ij}\}$ and $\Omega$ denotes the set of subscripts $(i,j)$ of the observed elements $y_{ij}$, $1 \leq i \leq N$, $1\leq j \leq M$. 
Unfortunately, accurately solving this problem is computationally difficult. One approach for 
practically overcoming this difficulty is to convert \eref{eq:rankmin}
 into a convex problem, which is solvable via semidefinite programming in a practical time scale 
by substituting the rank function with the nuclear norm \cite{Hastie15, CandesPlan10}. 
Surprisingly, this heuristic approach is known to fully recover the matrix itself 
under particular assumptions \cite{CandesPlan10,Keshavan10,Koltchinskii11}, making this 
treatment favorable for both practical use and theoretical analysis. 
Nevertheless, semidefinite programming that handles the full matrix explicitly 
encounters high computational costs in terms of both time and space complexity. \\
\indent One way to resolve this difficulty is to factorize the objective matrix, i.e., $X = U V^\ten$, and estimate the factor matrices $U,V$. 
Although this makes the problem non-convex,
 its global minimum is known to coincide with the optimal solution 
when $U$ and $V$ have rank higher than the true rank.  
Its desirable scalability, 
and the fact that all local minima equivalently minimize the nuclear norm
 given sufficient numbers of observations \cite{Ge16} make this formulation appealing for solving 
the matrix completion problem. 
\\
\indent Owing to the sparse nature of the problem, 
belief propagation (BP), or more commonly known as the cavity method in statistical physics, 
is an efficient method for finding the solution. The earliest application of BP-based algorithms to matrix completion 
is the message-passing version of alternating least squares (ALS-MP) \cite{KeshavanThesis}, 
which was rediscovered by \cite{Gamarnik16} as edge least squares. 
In a synthetic and noisy setup, the root-mean-squared error (RMSE) of the estimated matrix 
using ALS-MP was empirically demonstrated to be near an oracle bound. 
More recently, a cavity-based approach was proposed \cite{Noguchi19}, 
which had significantly lower time and space complexity, 
but required more observed entries compared with ALS-MP. 
More precisely, \cite{Noguchi19} demonstrated that the 
algorithm is as computationally efficient as the original version of alternating least 
squares \cite{Gamarnik16,Prateek13}, which is, however, outperformed by ALS-MP in terms of achieved performance. \\
\indent This study aims to develop another BP-based algorithm, 
which has the same time and space complexity as ALS-MP. 
The proposed method seeks to approximate message distributions of BP 
by Gaussians with moment matching imposed up to its second order,
which is analogous to the scheme used in expectation propagation (EP) \cite{Minka01}. 
Although the performance of our method is similar to that of ALS-MP under synthetic settings, 
experimental results show that our method is more robust in situations where the set of observed data 
is corrupted by non-Gaussian noise. 
Besides, we demonstrate that damping is necessary for the two algorithms to 
achieve optimal performance. This is confirmed by a comparison with the results from population dynamics (PD), 
which simulates the behavior of the message-passing algorithms in the large system limit. 
Approximate versions for both BP methods are also provided to reduce the necessary space complexity. 
Applications to real-world datasets indicate that our method outperforms conventional approaches. \\
\indent The outline of this paper is as follows.
In Section 2, we derive the BP-based algorithm and its memory-friendly version.
 In addition, we rederive ALS-MP in analog to our derivation. 
Comparisons between the two methods on an algorithmic level are also provided. 
Subsection 2.4 is devoted to explaining the PD algorithm,
 which is used to provide a crucial baseline for performance achievable in the large system limit. 
 The algorithm's performance via numerical experiments on synthetic and real datasets is presented in sections 3 and 4, respectively.
Finally, Section 5 summarizes this work and highlights possible future research directions. 
\section{The Algorithms}
By convex relaxation, \eref{eq:rankmin} is converted to the minimization of the nuclear norm as 
\begin{equation}\label{eq:nucnorm}
      \min_{X\in \mathbb{R}^{N \times M}} \norm{X}_\ast \quad {\rm subject\ to} \quad \abs{x_{ij} - y_{ij}}\leq \delta, \quad (i,j)\in \Omega,
\end{equation}
where $\norm{X}_\ast$ is the nuclear norm of matrix $X$, which is given by the sum of its singular values.
 Although this is readily solvable in polynomial time complexity 
via semidefinite programming, 
 we employ the equality \cite{Recht10} 
\begin{equation} \label{eq:nuctofro}
\norm{X}_{\ast} = \frac{1}{2}\min_{ \substack {U\in \mathbb{R}^{N \times R}, V\in \mathbb{R}^{M\times R}, \\ U V^\ten = X}} \norm{U}_F^2 + \norm{V}_F^2,     
\end{equation}
for $R \geq \rank X$ to further reduce its computational complexity. 
Here, $\norm{\cdot}_F$ denotes the Frobenius norm of the matrix. 
The Lagrangian dual of \eref{eq:nucnorm} using \eref{eq:nuctofro} is given by the following equation, where $\lambda$ is a 
function of $\delta$: \begin{equation} \label{eq:objective}
      \min_{U\in \mathbb{R}^{N \times R}, V\in \mathbb{R}^{M\times R}} \frac{1}{2} \sum_{(i,j)\in \Omega} \qty(y_{ij} - \vb{u}^\ten_i \vb{v}_j )^2 + \frac{\lambda}{2} \sum_{i = 1}^N \norm{\vb{u}_i}^2 +  \frac{\lambda}{2} \sum_{j = 1}^M \norm{\vb{v}_j}^2.      
\end{equation}
\indent \color{black}
Typically, $\lambda$ is a parameter controlling the strength of the nuclear 
norm regularization. 
Our introduction of $\lambda$ motivates us to tune this parameter 
to satisfy the constraint of \eref{eq:nucnorm}. However, the optimal 
value of $\delta$, and consequently $\lambda$, 
which offers the best performance 
is unknown in advance under most situations. 
In such cases, the parameter is to be determined using hyperparameter 
tuning techniques such as cross validation for best results.\\
\color{black}
\indent This optimization problem with respect to $N + M$ $R$-dimensional vectors 
$U = (\vb{u}_1, \cdots \vb{u}_N)^\ten, V = (\vb{v}_1, \cdots \vb{v}_M)^\ten$ 
is reduced to the dual form of \eref{eq:nucnorm}, which is $\min \sum_{(i,j)\in \Omega}(y_{ij}- x_{ij})^2 + \lambda \norm{X}_\ast$, 
if $R$ is larger than \color{black}or equal to \color{black} the rank of 
the optimal solution $X$. 
Throughout this paper, we focus on this factorized formulation. \\
\indent In other words, the solution is given by 
maximizing the posterior distribution that is composed of likelihood 
\begin{equation} \label{eq:likelihood}
      p( y_{ij}|\vb{u}_{i}, \vb{v}_j) \propto \exp \qty[- \frac{\beta}{2} (y_{ij} - \vb{u}_i^\ten \vb{v}_j)^2 ], \quad (i,j) \in \Omega,
\end{equation}
which implies that the observations are assumed to be corrupted by Gaussian noise with variance $\beta^{-1}$, 
and a Gaussian prior distribution that is dependent on the noise intensity:
\begin{equation}\label{eq:prior}
      p(\vb{u}_{i}) \propto \exp \qty( -\frac{\beta\lambda}{2}\norm{\vb{u}_i}^2 ), \quad p(\vb{v}_{i}) \propto \exp \qty( -\frac{\beta\lambda}{2}\norm{\vb{v}_j}^2 ), \quad 1\leq i\leq N, \ 1\leq j\leq M.
\end{equation}
\subsection{Derivation of the Gaussian BP algorithm}
\begin{figure}
      \raggedleft
      \includegraphics[width = 0.9\hsize]{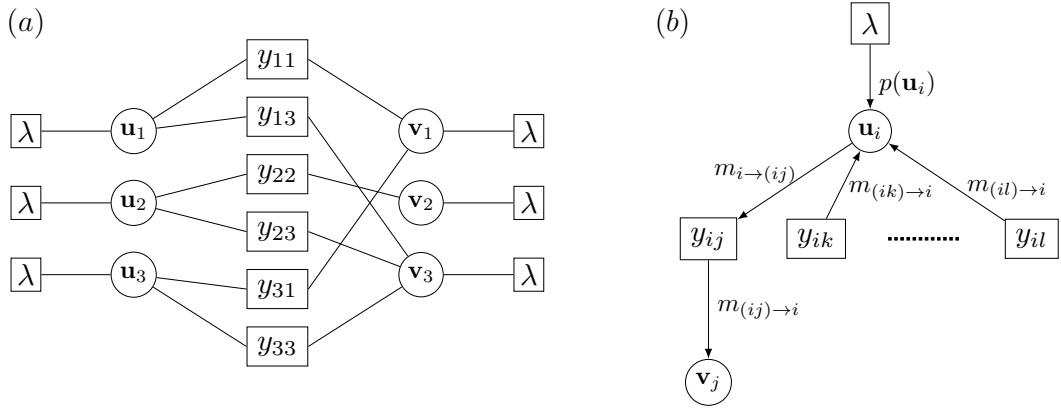}
      \caption{(a) An example of a factor graph for $N = M = 3$ and six observations. 
      (b) Illustration of the message-passing algorithm on a factor graph. Node $i$ collects 
      the prior message and the messages from $(ik), \ldots , (il)$, and passes $m_{i\to (ij)}$ to factor $(ij)$. 
      As a result, factor $(ij)$ passes message $m_{(ij)\to j}$ to node $j$. Irrelevant nodes and factors are not shown for ease of visualization. }
      \label{fig:factor}
  \end{figure}
The proposed BP algorithm aims at approximating the posterior distribution 
given by \eref{eq:likelihood} and \eref{eq:prior}. 
For this, we first express the variable dependence of the posterior on a factor graph (\fref{fig:factor}). 
Here, each interacting factor represents the likelihood for a single observed variable,
$p(y_{ij} | \vb{u}_i, \vb{v}_j)$, and is indexed by the pair of indices $(i,j)$. 
Without causing confusion, we denote the factors using the Greek letters $\mu, \nu, \ldots$ without explicitly 
writing the pair of indices. 
The prior factors, illustrated by the boxed $\lambda$, are individually connected to each variable node. \\
\indent The BP algorithm is an iterative procedure to find the fixed point solution of the following 
closed equations for the distributions defined on the edges of the factor graph:
 \numparts
 \begin{eqnarray}
       m^{t+1}_{\mu \rightarrow i}(\vb{u}_i) &\propto \int \dd \vb{v}_j \ p(y_{ij} | \vb{u}_i , \vb{v}_j) m^{t}_{j \rar \mu}(\vb{v}_j), \label{u_int}\\
       m^{t+1}_{i \rar \mu}(\vb{u}_i) &\propto p(\vb{u}_i)  \prod_{\nu \in \partial i \backslash \mu} m^{t+1}_{\nu \rar i}(\vb{u}_i), \label{u_sum} \\
       m^{t+1}_{\mu \rightarrow j}(\vb{v}_j) &\propto \int \dd \vb{u}_i \ p(y_{ij} | \vb{u}_i , \vb{v}_j) m^{t+1}_{i \rar \mu}(\vb{u}_i),\label{v_int} \\
       m^{t+1}_{j \rar \mu}(\vb{v}_j) &\propto p(\vb{v}_j)  \prod_{\nu \in \partial j \backslash \mu} m^{t+1}_{\nu \rar j}(\vb{v}_j). \label{v_sum}
      \end{eqnarray}
\endnumparts
The expression $\partial i$ denotes the set of factors connected to node $i$, whereas the superscript $t = 0,1,\ldots$ denotes the iteration number. 
The function $m^t_{\mu \rightarrow i}(\vb{u}_i)$ represents the marginal likelihood (or cavity bias) of $\vb{u}_i$, 
given by $y_{ij}$, whereas $m^t_{i \to \mu}(\vb{u}_i)$ is the marginal distribution of $\vb{u}_i$ in the absence of factor $y_{ij}$ (or the cavity distribution). 
The approximated posterior distributions of the nodes at iteration $t$ are given by 
\begin{eqnarray}
      p(\vb{u}_i | Y) \propto p(\vb{u}_i) \prod_{\mu \in \partial i} m^{t}_{\mu \rar i}(\vb{u}_i), \label{u_post} \\
      p(\vb{v}_j | Y) \propto p(\vb{v}_j)  \prod_{\mu \in \partial j} m^{t}_{\mu \rar j}(\vb{v}_j). \label{v_post}
\end{eqnarray} 
\indent Solving \eref{u_int}--\eref{v_sum} accurately
 is difficult, as these are functional equations with continuous degrees of freedom. 
Therefore, we approximately handle the BP equations using a few parameters. 
More explicitly, we parameterize the message distributions by those of the general $R$-dimensional Gaussian forms as
\numparts
\begin{eqnarray}
      m^{t}_{i \to \mu}(\vb{u}_i) &\propto \exp \qty[-\frac{\beta}{2} \vb{u}^\ten_i A_{i \to \mu}(t) \vb{u}_i +\beta \vb{B}_{i \to \mu}^\ten (t) \vb{u}_i ]\label{param_u_fn},\\
      m^{t}_{j \to \mu}(\vb{v}_j) &\propto \exp \qty[-\frac{\beta}{2} \vb{v}^\ten_j C_{j\to \mu}(t) \vb{v}_j +\beta \vb{D}_{j \to \mu}^\ten (t) \vb{v}_i ] \label{param_v_fn},
\end{eqnarray}
\endnumparts
where $A_{i\to \mu}, C_{j\to \mu} \in \mathbb{R}^{R\times R}$ and $\vb{B}_{i\to \mu}, \vb{D}_{j\to \mu} \in \mathbb{R}^{R}$ denote the 
natural parameters. 
\\
\indent For brevity, we hereafter omit the iteration number unless otherwise needed. 
Inserting \eref{param_v_fn} into the right-hand side of \eref{u_int}, we obtain 
the marginal likelihood as (a detailed derivation is given in the Appendix) 
\begin{equation}
      \label{tmp1}
     \int \dd \vb{v}_j \ p(y_{ij} | \vb{u}_i , \vb{v}_j) m_{j \rar \mu}(\vb{v}_j) \propto (1 + \vb{u}_i^\ten C_{j\to \mu}^{-1}\vb{u}_i)^{-1/2} \exp \qty[ -\frac{\beta}{2} \frac{(y_{ij} - \vb{u}_i^\ten C_{j\to \mu}^{-1} \vb{D}_{j\to \mu})^2}{1 + \vb{u}_i^\ten C_{j\to \mu}^{-1}\vb{u}_i} ].
\end{equation}
\color{black}
Note that if $\abs{\partial j}$ is sufficiently larger than $R$, the Gaussian parameterization is 
asymptotically exact. This is because the eigenvalue of matrix $C_{j\to \mu}$ typically 
scales with $\abs{\partial j}$ (see \eref{v_sum}), and thus $ \vb{u}_i^\ten C_{j\to \mu}^{-1}\vb{u}_i$ can 
be ignored. This does not necessarily imply that $\abs{\partial j}$ must be $O(N)$, 
and observations can be sparse, e.g. $\abs{\partial j} = O(\log N)$ for $R = O(1)$. 
The corresponding algorithm which applies this approximation is indeed ALS-MP, which is introduced in 
2.3. 
\color{black}
Otherwise, \eref{tmp1} is not of Gaussian form with respect to $\vb{u}_i$, 
and does not lead to a closed set of equations. 
To close the update equations for the parameters of $A$, $\vb{B}$, $C$, and $\vb{D}$ 
in the limit of $\beta \to \infty$, we resort to the moment matching condition up to second order similarily as employed in \cite{Minka01}.\\
\indent For this, we introduce the partition function based on \eref{tmp1} as 
\begin{equation} \label{tmp2}
      Z(\vb*{\theta}) = \int \dd \vb{u} \ (1 + \vb{u}^\ten C^{-1}\vb{u})^{-1/2} \exp \qty[ -\frac{\beta}{2} \frac{(y_{ij} - \vb{u}^\ten C^{-1} \vb{D})^2}{1 + \vb{u}^\ten C^{-1}\vb{u} } + \beta \vb*{\theta}^\ten \vb{u} - \frac{\beta \epsilon}{2}\norm{\vb{u}}^2 ],
\end{equation}
where we dropped all indices for notational simplicity and used a Gaussian factor $\exp\qty(-\beta \epsilon \norm{\vb{u}}^2/2), \ 0 < \epsilon \ll 1$, 
to prevent the integral from diverging. For $\beta \gg 1$, evaluating the right-hand side of \eref{tmp2} using the Laplace approximation yields 
an expression 
\begin{equation}
      \frac{1}{\beta}\log Z(\vb*{\theta}) = \max_{\vb{u}} \qty[  -\frac{1}{2} \frac{(y_{ij} - \vb{u}^\ten C^{-1} \vb{D})^2}{1 + \vb{u}^\ten C^{-1}\vb{u} } + \vb*{\theta}^\ten \vb{u} - \frac{\epsilon}{2}\norm{\vb{u}}^2 ],
\end{equation}
which offers the maximum condition as 
\begin{equation}
      \frac{y- \vb{u}^\ten \vb{v}^{\ast}}{1 + \vb{u}^\ten C^{-1}\vb{u}} \vb{v}^\ast +\qty(\frac{y- \vb{u}^\ten \vb{v}^{\ast}}{1 + \vb{u}^\ten C^{-1}\vb{u}})^2 C^{-1}\vb{u} + \vb*{\theta} - \epsilon \vb{u} = \vb{0},
\end{equation}
where $\vb{v}^\ast = C^{-1}\vb{D}$. For $\norm{\vb*{\theta}} \ll 1$ and $\epsilon \ll 1$, the solution to this equation is given by 
\begin{equation}
      \vb{u}^\ast(\vb*{\theta}) = \frac{y}{\norm{\vb{v}^\ast}^2}\vb{v}^\ast + \qty[ \epsilon \vb{1}_R + \frac{\vb{v}^\ast (\vb{v}^{\ast})^\ten}{1 + y^2 \alpha} ]^{-1} \vb*{\theta} + O(\norm{\vb*{\theta}}^2, \epsilon),
\end{equation}
where $\alpha \equiv (\vb{v}^\ast)^\ten C^{-1}\vb{v}^\ast / \norm{\vb{v}^\ast}^4$. 
This means that the first and second moments of the distribution derived by normalizing \eref{tmp1} are provided as 
\begin{equation}
      \expval{\vb{u}} = \frac{1}{\beta} \eval{\pdv{\vb*{\theta}}\log Z(\vb*{\theta})}_{\vb*{\theta} = \vb{0}} = \eval{\vb{u}^\ast(\vb*{\theta})}_{\vb*{\theta} = \vb{0}} = \frac{y}{\norm{\vb{v}^\ast}^2}\vb{v}^\ast + O(\epsilon)
\end{equation}
and 
\begin{equation}
      \expval{\vb{u}\vb{u}^\ten} - \expval{\vb{u}} \expval{\vb{u}}^\ten = \frac{1}{\beta^2} \eval{\pdv[2]{}{\vb*{\theta}}{\vb*{\theta}} \log Z(\vb*{\theta}) }_{\vb*{\theta} = \vb{0}} = \frac{1}{\beta} \eval{\pdv{\vb{u}^\ast (\vb*{\theta})}{\vb*{\theta}}}_{\vb*{\theta} = \vb{0}} = \frac{1}{\beta} \qty[ \epsilon \vb{1}_R + \frac{\vb{v}^\ast (\vb{v}^{\ast})^\ten}{1 + y^2 \alpha} ]^{-1} + o(\beta^{-1}),
\end{equation}
respectively, where $\vb{1}_R\in \mathbb{R}^{R\times R}$ is the identity matrix. 
Approximating the right-hand side of \eref{u_int} using a Gaussian function that reproduces these moments,
 which corresponds to the moment matching condition, and recovering the relevant indices results in 
\begin{equation} \label{closed_form_equation1}
      m_{\mu\to i}(\vb{u}) \propto \exp \qty(-\frac{\beta}{2} \frac{\vb{u}^\ten \vb{v}_{j\to \mu} \vb{v}^\ten_{j\to \mu} \vb{u}}{1 + y_{ij}^2 \alpha_{j\to \mu} } + \beta \frac{y_{ij}\vb{v}_{j\to \mu}^\ten \vb{u}}{1 + y_{ij}^2 \alpha_{j\to \mu}} ),
\end{equation}
under the limit $\epsilon \to +0$, 
where $\vb{v}_{j\to\mu} \equiv C_{j\to \mu}^{-1} \vb{D}_{j\to \mu}$ and $\alpha_{j\to \mu} \equiv \vb{v}_{j\to \mu}^\ten C_{j\to \mu}^{-1}\vb{v}_{j\to \mu} / \norm{\vb{v}_{j\to \mu}}^4$. 
Inserting this expression into \eref{u_sum} provides update rules for $A$ and $\vb{B}$ as
\begin{eqnarray}
      A_{i\to \mu} &= \lambda \vb{1}_{R} + \sum_{\nu \in \partial i \backslash \mu} \frac{ \vb{v}_{j \to \nu} \vb{v}_{j \to \nu}^\ten }{ 1 + y_{ij}^2\alpha_{j\to \nu} }, \label{GPBP1}\\
      \vb{B}_{i\to\mu} &=  \sum_{\nu \in \partial i \backslash \mu}  \frac{y_{ij} \vb{v}_{j\to \nu} }{ 1 + y_{ij}^2\alpha_{j\to \nu}}\label{GPBP2}.
\end{eqnarray}
Similarly, the update rules for $C$ and $\vb{D}$ are obtained as follows: 
\begin{eqnarray}
      C_{j\to \mu} &= \lambda \vb{1}_{R} + \sum_{\nu \in \partial j \backslash \mu} \frac{ \vb{u}_{i \to \nu} \vb{u}_{i \to \nu}^\ten }{ 1 + y_{ij}^2\alpha_{i\to \nu} }\label{GPBP3},\\
      \vb{D}_{j\to\mu} &=  \sum_{\nu \in \partial j \backslash \mu}  \frac{y_{ij} \vb{u}_{i \to \nu} }{ 1 + y_{ij}^2\alpha_{i \to \nu}}\label{GPBP4},
\end{eqnarray}
where $\vb{u}_{i\to\mu} \equiv A_{i\to \mu}^{-1}\vb{B}_{i\to \mu}$ and $\alpha_{i\to \mu} \equiv \vb{u}_{i\to \mu}^\ten A_{i\to \mu}^{-1}\vb{u}_{i\to \mu} / \norm{\vb{u}_{i\to \mu}}^4$.
The posterior distributions \eref{u_post} and \eref{v_post} are also given in Gaussian form as 
\begin{eqnarray}
      p(\vb{u}_i | Y) \propto \exp \qty( -\frac{\beta}{2} \vb{u}_i^\ten A_i \vb{u}_i +\beta \vb{B}_i^\ten \vb{u}_i ), \\
      A_i \equiv \lambda \vb{1}_{R} + \sum_{\mu \in \partial i} \frac{ \vb{v}_{j \to \mu} \vb{v}_{j \to \mu}^\ten}{ 1 + y_{ij}^2\alpha_{j\to \mu}} , \quad \vb{B}_i \equiv \sum_{\mu \in \partial i }  \frac{y_{ij} \vb{v}_{j\to \mu} }{ 1 + y_{ij}^2\alpha_{j\to \mu}}, \label{eq:sum}
\end{eqnarray}
and 
\begin{eqnarray}
      p(\vb{v}_j | Y) \propto \exp \qty( -\frac{\beta}{2} \vb{v}_j^\ten C_j \vb{v}_j +\beta \vb{D}_j^\ten \vb{v}_j ), \\
      C_j \equiv \lambda \vb{1}_{R} + \sum_{\mu \in \partial j} \frac{ \vb{u}_{i \to \mu} \vb{u}_{i \to \mu}^\ten}{ 1 + y_{ij}^2\alpha_{i\to \mu}} , \quad \vb{D}_j \equiv \sum_{\mu \in \partial j }  \frac{y_{ij} \vb{u}_{i\to \mu} }{ 1 + y_{ij}^2\alpha_{i\to \mu}}.
\end{eqnarray}
\\
\indent Three issues are noteworthy here. 
First, the derived algorithm is analogous to EP \cite{Minka01} 
in terms of requiring the Gaussians to yield the same first and second moments. 
However, unlike our algorithm, EP employs the moment matching condition for the joint 
distribution 
\[
p(\vb{u}_i, \vb{v}_j) \propto p(y_{ij}|\vb{u}_i, \vb{v}_j) m_{i\to \mu}(\vb{u}_i) m_{j \to \mu}(\vb{v}_j),      
\]
which, in the limit of $\beta \to \infty$, is reduced to a set of coupled nonlinear equations with respect to the moments of $\vb{u}_i$ and $\vb{v}_j$. 
Consequently, one cannot obtain closed forms of the update rules such as \eref{closed_form_equation1}--\eref{eq:sum}, 
which reduces the practicality of the algorithm. In contrast, our algorithm eliminates this difficulty by imposing the moment matching requirement 
on the cavity distributions \eref{u_int} and \eref{v_int}. 
 The second issue is regarding the computational cost. 
Although the algorithm involves the inverse of the matrix $A_{i\to \mu}$ and $C_{j\to\mu}$, these can be 
calculated explicitly using the Sherman--Morrison formula in $O(R^2)$ time complexity 
during the iterations once $A^{-1}_{i\to \mu}$ and $C^{-1}_{j\to\mu}$ are computed at the initial condition.
 Moreover, because the 
update equations depend only on the inverse matrices via $\alpha_{i\to \mu}, \alpha_{j\to \mu}$ 
or $\vb{u}_{i\to \mu}, \vb{v}_{i \to \mu}$, it is unnecessary to store these matrices in memory. 
This algorithm, therefore, has a space complexity of $O(\abs{\Omega}R)$. 
\color{black}
However, it should be noted the
 Sherman--Morrison formula is prone to high numerical errors compared 
 with standard matrix inversion when $\lambda$ is close to zero. 
 One must be cautious when employing the algorithm under such condition. 
\color{black}
Hereafter, we refer to this algorithm as the Gaussian \color{black}parameterized \color{black}belief propagation (GPBP) algorithm. 
The final issue is about a technique for improving the convergence property.
 For optimal performance, probabilistic damping \cite{Pretti05} can be employed in the algorithm, where the 
factor-to-node messages are updated using the weighted average of the new and old ones. 
In our algorithm, the damping procedure is given by
\begin{align}
      A_{i \to \mu}(t + 1) &= \lambda \vb{1}_{R} + \sum_{\nu \in \partial i \backslash \mu} \qty[(1-\gamma)\frac{ \vb{v}_{j \to \nu}(t) \vb{v}_{j \to \nu}^\ten(t) }{ 1 + y_{ij}^2\alpha_{j\to \nu}(t) }+ \gamma \frac{ \vb{v}_{j \to \nu}(t-1) \vb{v}_{j \to \nu}^\ten(t-1) }{ 1 + y_{ij}^2\alpha_{j\to \nu}(t-1) }], \\
      \vb{B}_{i \to \mu}(t + 1) &= \sum_{\nu \in \partial i \backslash \mu} \qty[ (1-\gamma) \frac{y_{ij} \vb{v}_{j\to \nu}(t) }{ 1 + y_{ij}^2\alpha_{j\to \nu}(t)} + \gamma \frac{y_{ij} \vb{v}_{j\to \nu}(t-1) }{ 1 + y_{ij}^2\alpha_{j\to \nu}(t-1)}  ]
\end{align}
for $\gamma \in [0,1]$. 
Our matrix is still a sum of a diagonal matrix and rank-one matrices, which indicates that 
the  time complexity of the algorithm is nonetheless $O(\abs{\Omega}R^2)$. 
Note that $A_{i \to \mu}(t + 1)$ and $\vb{B}_{i \to \mu}(t + 1)$ 
are not strictly the weighted sums of $A_{i \to \mu}(t)$ and $\vb{B}_{i \to \mu}(t)$, but their updated values. 
In fact, our damping method is equivalent to that in \cite{Pretti05} only when $\gamma$ is sufficiently small.
However, at least for the experiments conducted in this study, the employed values provided satisfactory results. 
\subsection{Approximate BP algorithm}
The space complexity of GPBP is $O(\abs{\Omega}R)$, which may be computationally 
intense when $\abs{\Omega}$ is large. 
To relax this bottleneck, we apply a common approximation used in deriving 
approximate message passing (AMP) 
algorithms from primary BP algorithms \cite{Kabashima03}.
 This scheme exploits the fact that each node-to-factor message 
 differs slightly from the sum of the factor-to-node messages when the degree per node is sufficiently large. \\
\indent Using the Sherman--Morrison formula, the cavity vector at sweep iteration $t$, $\vb{v}_{j\to \mu}(t)$, is evaluated as 
\begin{align}
      \vb{v}_{j\to \mu}(t) &= C_{j\to \mu}^{-1}(t) \vb{D}_{j\to \mu}(t) \\
      & = \qty[C_j^{-1}(t) + \frac{C_j(t)^{-1} \vb{u}_{i\to\mu}(t) \vb{u}_{i\to\mu}^\ten (t) C_j^{-1}(t)}{1 + y_{ij}^2\alpha_{i\to\mu}(t) - \vb{u}_{i\to\mu}^\ten (t) C_j^{-1}(t) \vb{u}_{i\to\mu}(t)}]\qty[\vb{D}_j(t) - \frac{y_{ij}\vb{u}_{i\to \mu}(t)}{1 + y_{ij}^2 \alpha_{i\to\mu}(t)}]\\
      & \simeq \vb{v}_j(t) - \frac{y_{ij} - \vb{u}_i^\ten (t) \vb{v}_j(t)}{1 + y_{ij}^2 \alpha_i(t) - \vb{u}^\ten_i(t) C_j^{-1}(t) \vb{u}_i(t)}C^{-1}_j(t) \vb{u}_i(t)\equiv \tilde{\vb{v}}_{j\to \mu} (t)  , \label{v_cav_approx}
\end{align}
where we used the zeroth approximation $\vb{u}_{i\to \mu}(t) \simeq \vb{u}_i(t)$ to derive the third line, and 
$\alpha_i(t) \equiv \vb{v}_j^\ten (t) C_j^{-1}(t) \vb{v}_j(t) / \norm{\vb{v}_j(t)}^4$. Similarly, 
\begin{equation}
      C^{-1}_{j\to \mu}(t) \simeq C_j^{-1}(t) + \frac{C_j^{-1}(t) \vb{u}_i(t)\vb{u}_i^\ten (t) C_j^{-1}(t) }{1 + y_{ij}^2 \alpha_i(t) - \vb{u}_i^\ten (t) C_j^{-1}(t)\vb{u}_i(t)} \equiv \tilde{C}^{-1}_{j\to \mu}(t),  \label{C_cav_approx}
\end{equation}
\begin{equation}
      \alpha_{j\to \mu}(t) \simeq \frac{\tilde{\vb{v}}^\ten_{j\to \mu}(t)\tilde{C}^{-1}_{j\to \mu}(t)\tilde{\vb{v}}_{j\to \mu}(t) }{ \norm{\tilde{\vb{v}}_{j\to \mu}(t)}^4 } \equiv \tilde{\alpha}_{j\to \mu} (t) .\label{alpha_cav_approx}
\end{equation}
Substituting \eref{v_cav_approx}--\eref{alpha_cav_approx} into \eref{eq:sum}, we obtain the approximated form of 
$A_{i}(t + 1)$ and $\vb{u}_i(t + 1)$. Similar update equations are obtained for $C_j(t + 1)$ and $\vb{v}_j(t + 1)$.
\\
\indent It is crucial to correctly evaluate the time dependency of the variables 
when deriving the approximate algorithms. Although the approximations lead to an iterative algorithm 
solving a Thouless--Anderson--Palmer-like equation \cite{TAP77} for our system, earlier 
research \cite{Manoel15} showed that intuitive iteration schemes exhibit suboptimal convergence properties. 
In the above derivation, the time dependencies are completely analogous to GPBP. 
This approach of mimicking BP was also taken in \cite{Manoel15} for compressed sensing, 
and empirically showed better convergence compared with other methods. 
To the best of our knowledge, this scheme also gave the best results in our case. 
\\
\indent Compared with GPBP, damping has a significant influence on the resultant performance in the approximate algorithm. 
Approximate probability damping is employed similarly to the case of GPBP 
by directly damping the node variables using \eref{v_cav_approx} and \eref{alpha_cav_approx} as
\begin{align}
      A_{i}(t + 1) &=  \lambda \vb{1}_{R}+ \sum_{\mu \in \partial i} \qty[ (1-\gamma) \frac{ \tilde{\vb{v}}_{j\to \mu}(t) \tilde{\vb{v}}_{j\to \mu}^\ten(t) }{1 + y_{ij}^2 \tilde{\alpha}_{j\to \mu}(t)} + \gamma \frac{ \tilde{\vb{v}}_{j\to \mu}(t-1) \tilde{\vb{v}}_{j\to \mu}^\ten(t-1) }{1 + y_{ij}^2 \tilde{\alpha}_{j\to \mu}(t-1)}  ]  , \\
       \vb{B}_{i}(t + 1) &=  \sum_{\mu \in \partial i}  \qty[ (1-\gamma) \frac{y_{ij} \tilde{\vb{v}}_{j\to \mu}(t) }{ 1 + y_{ij}^2\tilde{\alpha}_{j\to \mu}(t)} + \gamma \frac{y_{ij} \tilde{\vb{v}}_{j\to \mu}(t-1) }{ 1 + y_{ij}^2\tilde{\alpha}_{j\to \mu}(t-1)} ] .
\end{align}
Although this implementation requires storing the matrix parameters $A_i, C_i$ explicitly, only the node variables 
must be kept in memory, which leads to a reduction in the space complexity from $O(\abs{\Omega}R)$ to $O(\abs{\Omega} + (N + M)R^2)$. 
As $\abs{\Omega} > (N + M)R^2$ must hold for 
making a matrix recoverable, this reduction in space complexity is 
beneficial, especially in cases where the number of observations $\abs{\Omega}$ is larger than $N + M$.
 We refer to this algorithm as the approximated Gaussian \color{black}parameterized \color{black} belief propagation (approxGPBP) algorithm.
\subsection{Relation to ALS-MP}
Unlike GPBP, which handles matrices and vectors,
 ALS-MP is an iterative algorithm for manipulating only vectors defined on the edges of the factor graphs. 
However, ALS-MP can be derived from the BP framework with a few modifications from the derivation of GPBP.\\
\indent Given factor-to-node messages $m_{\mu \to i}(\vb{u}_i)$ and $m_{\mu \to j}(\vb{v}_j)$, we can define 
{\it cavity vectors} by maximizing the cavity distributions as 
\begin{eqnarray}
      \vb{u}_{i\to \mu} = \mathop{\rm argmax}_{\vb{u}_i} p(\vb{u}_i) \prod_{\nu \in \partial i \backslash \mu} m_{\nu \to i}(\vb{u}_i),\label{eq:alsmp1}\\
      \vb{v}_{j\to \mu} = \mathop{\rm argmax}_{\vb{v}_j} p(\vb{v}_j) \prod_{\nu \in \partial j \backslash \mu} m_{\nu \to j}(\vb{v}_j)\label{eq:alsmp2}.
\end{eqnarray} 
Instead of computing the marginal likelihoods in \eref{u_int} and \eref{v_int}, ALS-MP evaluates 
factor-to-node messages by inserting the cavity vectors $\vb{v}_{j\to \mu}$ and $\vb{u}_{i\to \mu}$ 
into the likelihood function as 
\begin{eqnarray}
      m_{\mu \to i}(\vb{u}_i) \propto p(y_{ij} | \vb{u}_i, \vb{v}_{j\to \mu}), \label{eq:alsmp3}\\
      m_{\mu \to i}(\vb{v}_j) \propto p(y_{ij} | \vb{u}_{i\to \mu}, \vb{v}_j ) \label{eq:alsmp4}.
\end{eqnarray} 
By assigning update indices appropriately, \eref{eq:alsmp1}--\eref{eq:alsmp4} lead to the following update equations for $\vb{u}_{i\to \mu}$ and $\vb{v}_{j\to \mu}$:
\begin{eqnarray} 
      \vb{u}_{i\to \mu} (t + 1) &= \qty( \lambda \vb{1}_{R} + \sum_{\nu \in \partial i \backslash \mu} \vb{v}_{j\to \nu} (t) \vb{v}_{j \to \nu}^\ten (t) )^{-1} \qty(\sum_{\nu \in \partial i \backslash \mu} y_{ij} \vb{v}_{j\to \nu}(t)), \label{alsmp_u}\\
      \vb{v}_{j\to \mu} (t + 1) &= \qty( \lambda \vb{1}_{R} + \sum_{\nu \in \partial i \backslash \mu} \vb{u}_{i\to \nu} (t+1) \vb{u}_{i \to \nu}^\ten (t+1) )^{-1} \qty(\sum_{\nu \in \partial j \backslash \mu} y_{ij} \vb{u}_{i\to \nu}(t+1)).\label{alsmp_v}
\end{eqnarray} 
These are somewhat similar to \eref{GPBP1}--\eref{GPBP4}. 
Indeed, GPBP is reduced to ALS-MP by dropping all $\alpha$ coefficients. 
Recall that $\alpha_{i\to \mu}$ and $\alpha_{j\to \mu}$ are defined as $\alpha_{i\to \mu} = \vb{v}_{j\to\mu}^\ten C_{j\to \mu}^{-1} \vb{v}_{j\to \mu} / \norm{\vb{v}_{j\to \mu}}^4$
and $\alpha_{j\to \mu} = \vb{u}_{i\to\mu}^\ten A_{i\to \mu}^{-1} \vb{u}_{i\to \mu} / \norm{\vb{u}_{i\to \mu}}^4$.
As $C_{j\to \mu}^{-1}$ and $A_{i\to \mu}^{-1}$ are proportional to
 $R\times R$ Fisher information matrices for cavity distributions, this implies that 
 $\alpha_{i\to \mu}$ and $\alpha_{j\to \mu}$ play the role of controlling the effect of 
 the observations depending on the uncertainty 
 of the cavity vectors. This property may be beneficial for making estimates robust. 
\\
\indent As for its computational aspect, ALS-MP has the same time and space complexity as GPBP, 
and the same damping techniques can be employed. 
In addition, its space-saved version, approxALS-MP, can be derived similarly to approxGPBP. 

\subsection{Performance Evaluation via Population Dynamics} \label{subsection:PD}
The performance of GPBP and ALS-MP can be evaluated by carrying out experiments for many random instances 
of the matrices under the given conditions. 
However, PD \cite{MM09}, which is a sampling method, 
is more efficient for examining the typical performance of the algorithms in the large system limit of $N, M\to \infty$. 
\color{black}
In fact, PD can also be regarded as a method for finding the replica symmetric solution for this system under the framework of the replica theory \cite{MM09,Nishimori01}, 
while applying the approximations which GPBP and ALS-MP adopt. 
Interested readers may refer to Appendix B for a derivation of the replica symmetric solution 
and further discussions about its relation with PD. 
\color{black}
\\
\indent We focus on PD for GPBP performed in cases where each row of 
matrices $U$ and $V$ are linked randomly with $c$ and $r$ observations in $Y$, 
respectively, but its generalization to ALS-MP is straightforward. 
For this, we prepare $N_{\rm PD}\gg 1$ tuples of $\vb{u}^0 \in \mathbb{R}^R$, 
$\vb{u}_{\rm cav} \in \mathbb{R}^R$, and $\alpha_u \in \mathbb{R}$
for the estimate of $U$, which are stored in a reservoir that we term ``$U$-pool.'' 
In addition, their counterparts $\vb{v}^0$, $\vb{v}_{\rm cav}$, and $\alpha_v$ 
are also prepared in ``$V$-pool'' for the estimation of $V$. 
Here, $\vb{u}^0$ and $\vb{v}^0$ correspond to the transposes of row vectors 
of the true matrices $U^0$ and $V^0$, whereas $\vb{u}_{\rm cav}$, 
$\alpha_u$, $\vb{v}_{\rm cav}$, and $\alpha_v$ represent instances of messages. 
To update each message tuple in the $U$-pool, we randomly select $c-1$ tuples of 
$\vb{v}^0$, $\vb{v}_{\rm cav}$, and $\alpha_v$ from the $V$-pool, 
and renew $\vb{u}_{\rm cav}$ and $\alpha_u$ following the GPBP algorithm 
handling the $c-1$ $\vb{v}_{\rm cav}$ and $\alpha_v$ as
$\vb{v}_{j\to \nu}$ and $\alpha_{j\to \nu}$ and setting 
$y_{ij} = (\vb{u}^0)^\ten \vb{v}^0 + z$, where $z$ is an independent sample 
from a certain distribution with zero mean. 
Similar updates are performed for $\vb{v}_{\rm cav}$ and $\alpha_v$
using $r-1$ tuples of $\vb{u}^0$, $\vb{u}_{\rm cav}$, and $\alpha_u$, which are 
randomly chosen from the $U$-pool. 
After iterating these procedures many times, 
the populations of the message tuples converge to stationary distributions. 
Then, the estimate of $\vb{u}^0$ is computed by $c$ message tuples chosen randomly 
from the $V$-pool, and similarly for $\vb{v}^0$ by $r$ message tuples from the $U$-pool. 
\\
\indent PD simulates the macroscopic behavior of BP when influences of cycles in variable dependence 
are ignored. The typical lengths of the cycles tend to infinity as $N, M \to \infty$, and their 
influences asymptotically vanish when observations in $Y$ are linked randomly with rows of $U$ and $V$. 
Therefore, if messages of BP converge to a fixed point, the performance evaluated by the corresponding PD 
can be regarded as that achieved in the large system limit. 
However, there is a possibility that the 
messages will continue to move microscopically, even if their distributions converge macroscopically. 
In such cases, considerable deviations can be observed between the results of the direct BP experiments  
and the predictions by PD, which is related to the notion of replica symmetry breaking 
\cite{Kabashima03,Takahashi20}.

\section{Numerical Experiments: Synthetic datasets}
\subsection{Comparison of GPBP and ALS-MP}
\begin{figure}
      \raggedleft
       \includegraphics[width = 0.9\hsize]{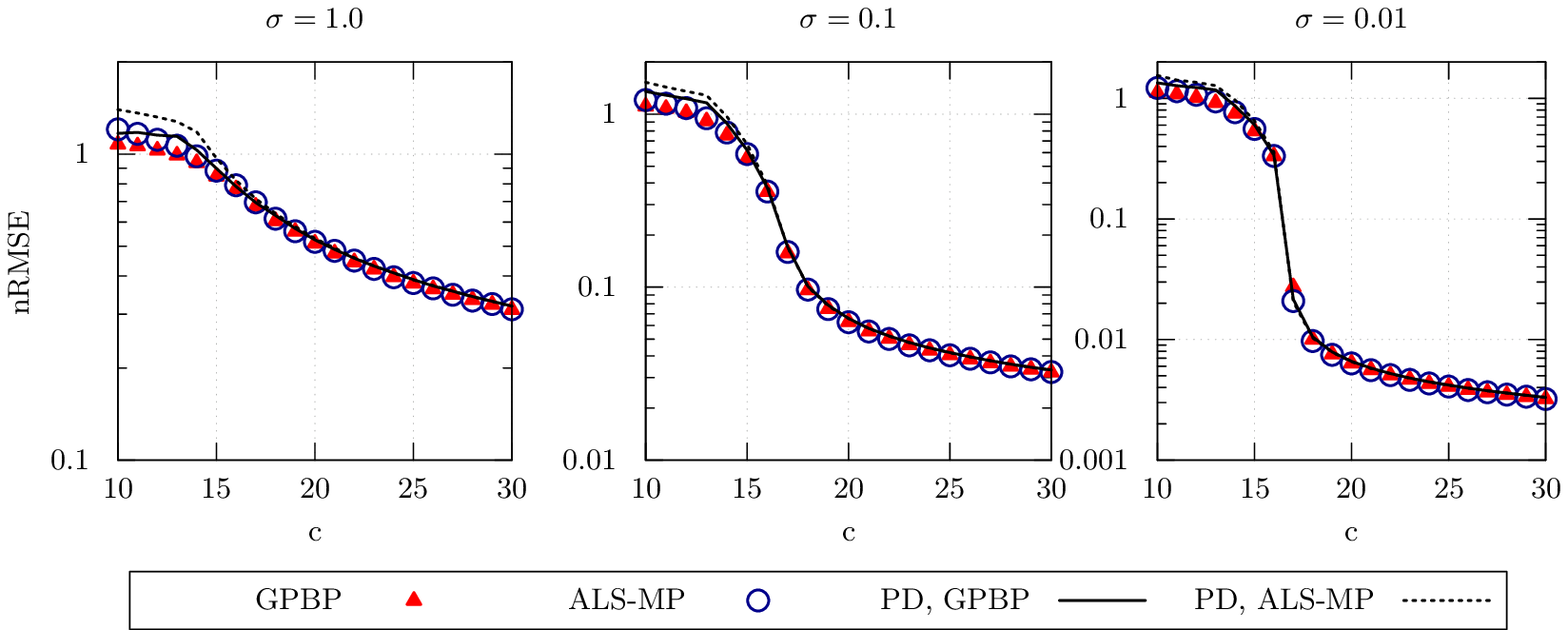}
       \caption{nRMSE of GPBP and ALS-MP for the Gaussian noise case. 
       Throughout the three noise intensities, GPBP and ALS-MP are in agreement when $c$ is above a transitional point. 
       Markers were obtained from the average of 10 random instances. }
      \label{fig:pop}
      \raggedleft
      \includegraphics[width = 0.9\hsize]{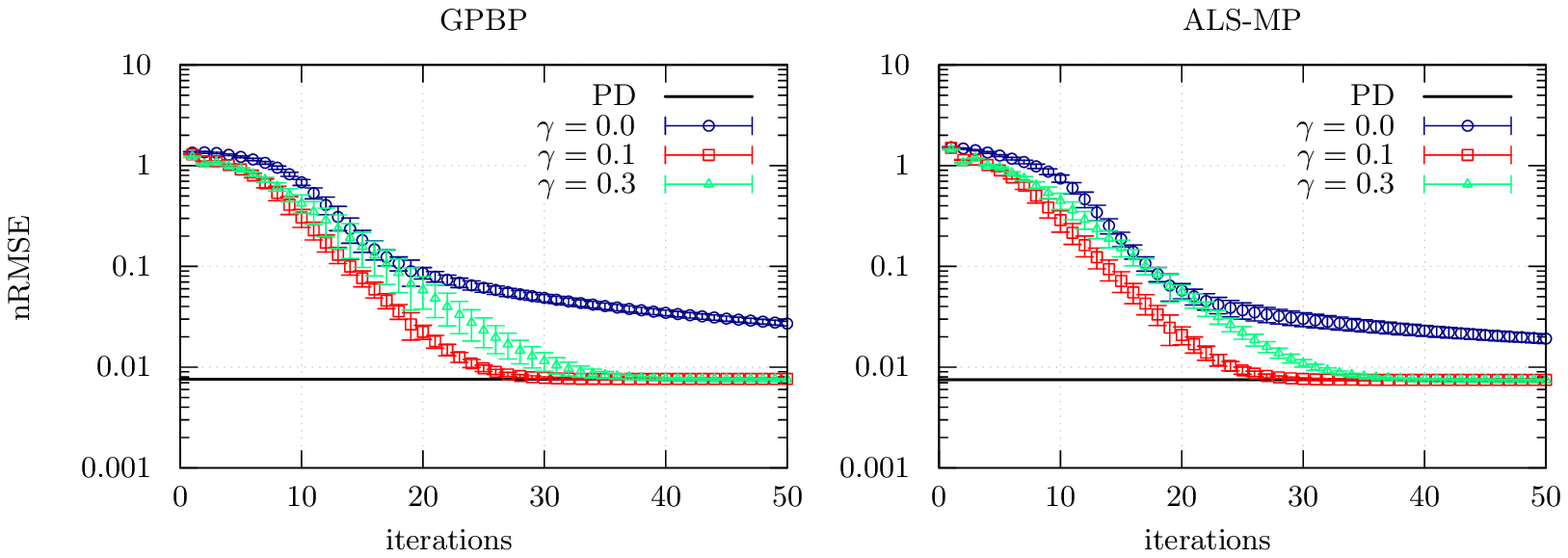}
      \caption{Convergence of GPBP and ALS-MP for $c = 19$ and $\sigma = 0.01$. Both algorithms exhibit a slow convergence to 
      the theoretical prediction given by PD when damping is absent. 
      Markers and error bars represent the averages and standard errors evaluated from 10 random instances.}
      \label{fig:damptraj}
  \end{figure}
\begin{figure}
      \raggedleft
      \includegraphics[width = 0.9\hsize]{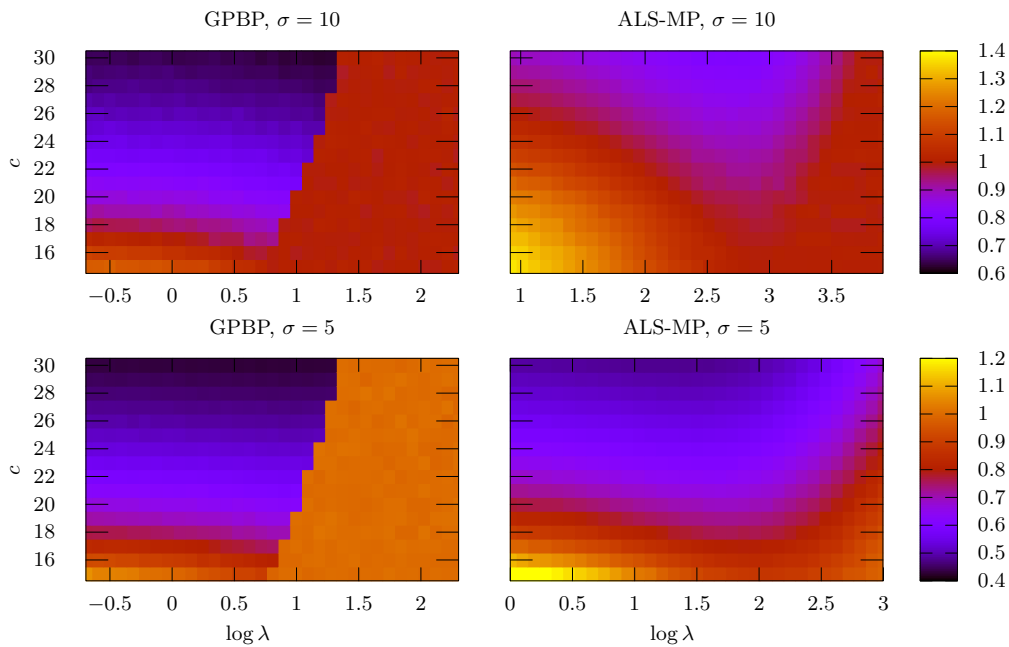}
      \caption{nRMSE of GPBP and ALS-MP assessed by PD.
       For both $\sigma = 5$ and $10$, 
      GPBP achieves a lower nRMSE when $\lambda$ is set to be optimal for each $c$. 
      Abrupt performance deterioration occurs for GPBP. }
      \label{fig:heatmap}
  \end{figure}
\begin{figure}
      \raggedleft
      \includegraphics[width = 0.9\hsize]{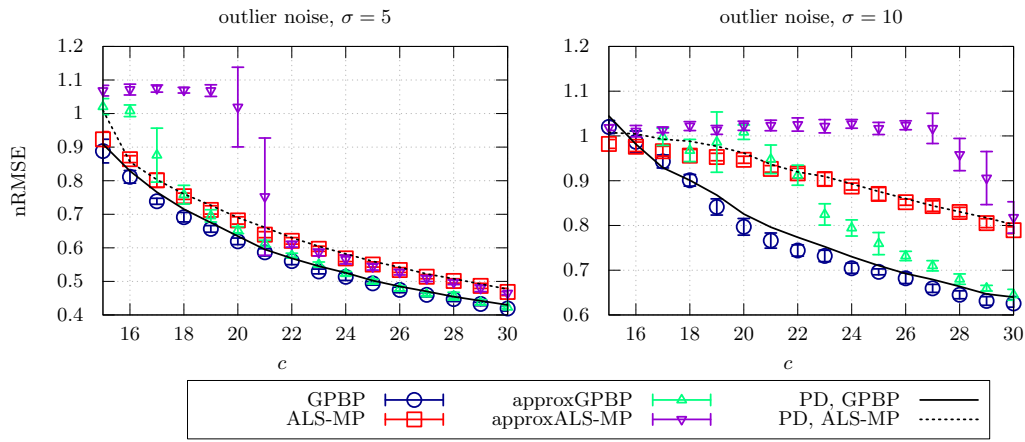}
      \caption{nRMSE of GPBP, ALS-MP, approxGPBP, and approxALS-MP near the optimal value of $\lambda$ ($\lambda_{\rm GPBP} = 1.85, \lambda_{\rm ALS-MP} = 4.91$ 
      for $\sigma = 5$, and $\lambda_{\rm GPBP} = 1.85, \lambda_{\rm ALS-MP} = 14.8$ 
      for $\sigma = 10$). Markers and error bars 
      represent averages and standard errors evaluated from 10 random instances.}
      \label{fig:cut}
  \end{figure}

Synthetic numerical experiments were conducted for both algorithms and their PD counterparts to 
investigate their performance. The dataset was prepared by generating the original 
uncorrupted matrix $X\in \mathbb{R}^{N\times M}$ from 
$U^0\in \mathbb{R}^{N\times R}, V^0\in \mathbb{R}^{M\times R}$, where the entries 
of $U^0, V^0$ are sampled independently from a standard Gaussian distribution $\mathcal{N}(0, 1)$. 
Throughout the experiments, the values of $M / N$ and $R$ were fixed to $M / N = 2$ 
and $R = 10$, where $N =500$ for both GPBP and ALS-MP, 
whereas $N_{\rm PD} = 2000$ for PD experiments. 
For PD, only the results from a single instance of $U^0$ and $V^0$ were obtained 
because we can expect the self-averaging property to hold. \\
\indent The observation $Y$ is given by $Y = U^0(V^0)^\ten + Z$, where $Z\in \mathbb{R}^{N\times M}$
is a noise matrix. 
Two types of noise corruptions were considered: Gaussian noise, 
where the entries of $Z$ are sampled independently from a Gaussian distribution $\mathcal{N}(0, \sigma^2)$
 and sparse noise, where the entries of $Z$ are sampled independently from a 
 Bernoulli--Gaussian distribution $0.9 \delta(z) + 0.1\mathcal{N}(0, \sigma^2)$. 
In the Gaussian noise setup, the regularization parameter was set to $\lambda = \sigma^2$, 
which corresponds to the cases where the problem of \eref{eq:objective} intends to maximize 
the correct posterior distribution composed of \eref{eq:likelihood} and \eref{eq:prior}. 
On the other hand, the sparse noise setup represents a situation where the minimization of the squared error 
$(y_{ij}- \vb{u}_i^\ten \vb{v})^2$ does not match the likelihood maximization, which may be more 
realistic than the Gaussian noise setup. 
For both cases, the entries were observed such that each row or column has the same number of 
observations. Therefore, $c\in \mathbb{N}$ observations were made for each column of matrix $Y$, whereas 
$r = c M/N \in \mathbb{N}$ observations were made for each row. 
\\
\indent The average performance of GPBP and ALS-MP was evaluated via the normalized RMSE (nRMSE)
 \begin{equation}
     {\rm nRMSE} = \sqrt{(NMR)^{-1}\sum_{i  =1}^N \sum_{j = 1}^M \qty(x_{ij} - \vb{u}_i^\ten \vb{v}_j)^2 }.
 \end{equation}
\Fref{fig:pop} shows nRMSE obtained by GPBP and ALS-MP in the Gaussian noise setup. 
The behavior of the two algorithms is similar. Differences in the performance, 
particularly at large values of $c$ and small values of $\sigma$, 
are unclear. It is worth noting that undamped 
GPBP and ALS-MP cannot achieve the theoretical values predicted by PD,  
particularly in cases where $\lambda$ and $\sigma$ are small.  
This is evident from \fref{fig:damptraj}, where 
the non-damped cases for both GPBP and ALS-MP show worse performance  
than even the slightly damped case ($\gamma = 0.1$). 
This indicates that damping is a crucial device for employing BP for the current problem.\\
\indent The sparse noise setup highlights the difference between GPBP and ALS-MP.
\Fref{fig:heatmap} shows the nRMSE given by PD for different values of $\lambda$. For
 $\sigma = 5$ and $10$, GPBP outperforms ALS-MP
at their optimal value of $\lambda$. The optimal value of $\lambda$ for
ALS-MP is also significantly larger than that for GPBP, indicating that ALS-MP 
is more likely to overfit and, hence, is more vulnerable to noise of non-Gaussian types. 
As stated in Section 2.3, this may be a result of GPBP being able to control the effect of observations more precisely. 
\Fref{fig:cut} shows nRMSE given by PD and algorithmic results when $\lambda$ is fixed near its
optimal value. Although deviations between PD and algorithmic results are evident, particularly in
large values of $\sigma$, overall the two results match fairly well. 
Interestingly, the qualitative behavior of GPBP and ALS-MP differs greatly with respect to different values of $\lambda$.
Although an abrupt performance deterioration occurs for GPBP above a certain value of $\lambda$,
 where the solution discontinuously shrinks to $U = 0$, $V = 0$, the nRMSE of ALS-MP varies smoothly. 
This implies the possibility that GPBP exhibits a discontinuous phase transition, whereas that of ALS-MP is 
continuous. 
\subsection{Comparison of approxGPBP and approxALS-MP}
\begin{figure}
    \raggedleft
    \includegraphics[width = 0.9\hsize]{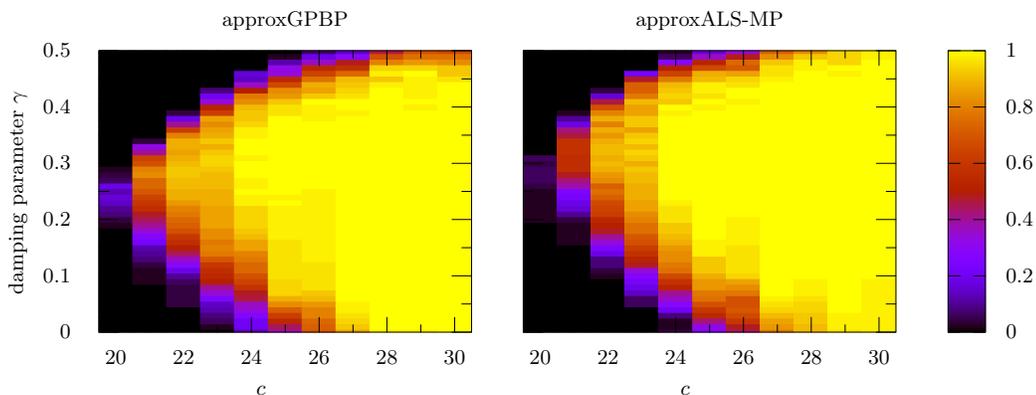}
    \caption{Reconstruction rate of approxGPBP and approxALS-MP for $\sigma = \epsilon = 0.01$ obtained from 
    100 random instances under different damping parameters.
    Both algorithms can reconstruct the matrix from $c \sim 22$ under appropriate damping. }
    \label{fig:dampswp}
\end{figure}
Damping plays a more significant role to achieve optimal 
performance for approxGPBP and approxALS-MP. 
In fact, both approximation algorithms sometimes cease to converge when $c$ is small. 
Therefore, we evaluate the performance of the algorithms by 
reconstruction rate, defined by the empirical probability of nRMSE being lower than 
some threshold value $\epsilon$. \\
\indent \Fref{fig:dampswp} shows the reconstruction rate of approxGPBP and approxALS-MP 
for Gaussian noise with $\sigma = \epsilon = 0.01$ for different damping parameters. 
Tuning the damping parameters appropriately significantly improves the reconstruction quality; 
the reconstruction threshold can be reduced from $c \sim 26$ to $c \sim 22$, 
when damping is optimal. 
Quantitatively, similar results were obtained for different values of $\sigma$ and $\epsilon$. 
Both algorithms exhibit similar performance (in terms of the threshold and nRMSE) in the Gaussian noise setup, 
which is consistent with the results obtained for the non-approximated counterparts. \\
\indent Consistency with the results from the non-approximated algorithms 
is also evident in the sparse noise setup in \fref{fig:cut}, 
where approxGPBP outperforms approxALS-MP in both noise intensities. 
The performance of approxGPBP and approxALS-MP seems to encounter a
transition from an uninformative 
to informative phase, with a margin where approxGPBP can obtain information on the 
matrix ($\rm nRMSE<1$), whereas approxALS-MP cannot. 
We speculate that this is another benefit gained by employing
 more informative messages in GPBP/approxGPBP. 
Nevertheless, as $c$ increases, the performance of approxGPBP and approxALS-MP 
asymptotically approaches that of the non-approximated versions, 
indicating that our perturbation treatment is valid. 
\section{Numerical Experiments: Real-world Datasets}
\begin{figure}
      \raggedleft
      \includegraphics[width = 0.9\hsize]{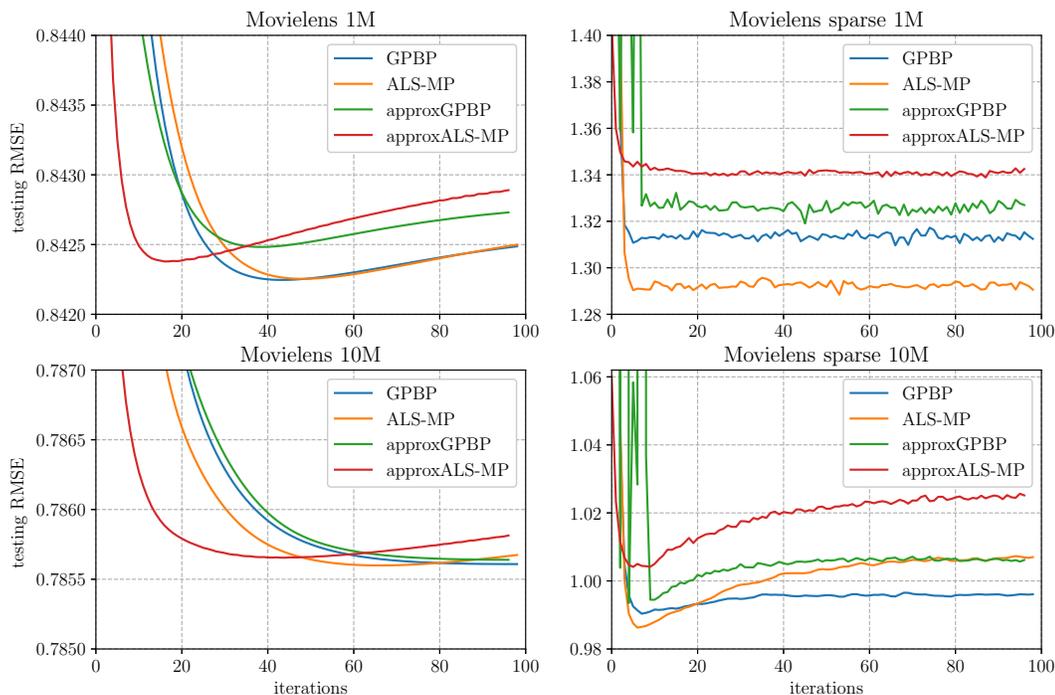}
      \caption{RMSE of the four algorithms for the (sparse) 1M and (sparse) 10M dataset. 
      The RMSE was obtained from the average of a 10-fold cross-validation.}
      \label{fig:movielens}
  \end{figure}
The performance of the algorithms in practical settings was evaluated via 
application to the 1M and 10M Movielens datasets \cite{movielens}, which are commonly used benchmark datasets for recommender systems. 
The 1M (10M) dataset consists of 1000209 (10000054) discrete ratings given by 6040 (69878) users 
on 3952 (10677) movies. All users are guaranteed to have rated at least 20 movies. 
The average number of ratings per user/movie is $c \sim 100$, which may 
make the task relatively easy owing to its dense connection. Therefore, a sparsified version of 
1M and 10M Movielens datasets, which only takes into account users who gave less than 31 ratings, 
was used as the benchmark. This subset, denoted as the sparse 1M (10M) Movielens dataset, 
has 18169 (295831) ratings given by 750 (12343) users on 2356 (6484) movies.\\
\color{black}
\indent We evaluated the algorithm's performance based on a 10-fold nested cross-validation procedure;
 each dataset was split into 10 random subsets of equal size, 
 and 9 out of 10 subsets were used for training, 
 while the remaining dataset was used to assess the RMSE. 
Five percent of the training dataset was held out as a validation dataset to determine the 
value of $\lambda$, while the remaining 95\% was used for training. 
The average of the 10 RMSE values was reported as the final RMSE score. 
The rank of the matrix $U, V$ was set to $R = 10$. The regularization parameter was 
chosen from 11 geometrically spaced values in the range $\qty[1, 5]$ for 1M, sparse 1M, and sparse 10M datasets, and 
6 geometrically spaced values in range $\qty[1, 5]$ for the 10M dataset. \\
\color{black}
\indent Results \color{black} from the above procedure \color{black} 
on the four datasets are given in \fref{fig:movielens}. 
As speculated, 
GPBP and ALS-MP exhibit little difference in performance on the relatively dense 1M dataset. The 10M dataset shows 
a similar trend, although ALS-MP converges faster than GPBP. However, ALS-MP presents 
signs of overfitting, where the test RMSE increases after achieving a minimum value. 
Mixed results are obtained for sparse datasets. Although dominant for the sparse 1M dataset, 
ALS-MP strongly overfits the training data for the sparse 10M dataset. 
Consistently, approxGPBP outperforms approxALS-MP for all four datasets, 
suggesting that our Gaussian treatment of messages is also beneficial for handling real-world data with low 
space complexity. 
\section{Conclusion}
In this study, we developed a Gaussian-based BP algorithm, 
GPBP, for the noisy matrix completion problem. 
By factorizing the inferred matrix and parameterizing the cavity distributions as Gaussians, 
the problem of continuous degrees of freedom in message passing was reduced to that of only a few 
variables.
The relation to a similar message-passing algorithm, ALS-MP, developed in the literature, was discussed. 
In addition, approximate but memory-friendly versions of GPBP and ALS-MP, namely approxGPBP and approxALS-MP, respectively,
 were derived by a perturbation treatment. \\
 \indent Experiments on synthetic data with Gaussian noise indicated that there is little to no difference in performance 
 between GPBP and ALS-MP. In contrast, those with non-Gaussian noise showed 
 that GPBP can exhibit better performance compared with ALS-MP. 
 A similar result was also obtained for their approximate counterparts, approxGPBP and approxALS-MP. \\
\indent Experiments on the Movielens datasets indicated that GPBP and ALS-MP provide similar performance, but 
GPBP is more robust against overfitting under fewer data. 
The experiments also showed that approxGPBP exhibits better performance than approxALS-MP for the datasets. 
This implies that for larger datasets where space complexity is an issue, 
approxGPBP could be a better choice. \\
\indent It is important to emphasize that although the Gaussian approximation of the posterior distribution 
empirically offered good performance, theoretically, this is suboptimal. This is because the matrix factorization model
 belongs to a family of singular statistical models \cite{Watanabe01}, where the 
 posterior distribution is generally approximated poorly by a Gaussian distribution even when many data are employed.
 It is important to examine the effect of singularity on factorized Gaussian approximations, 
and whether further improvements can be performed by a more legitimate parameterization. \\
 \indent Although our approach is based on nuclear norm minimization, several recent studies (for example, \cite{Salakh08}) 
 achieved a lower error in real-world datasets in more complex problem settings. However, 
our method can still be advantageous in terms of time or space complexity owing to its simplicity. 
Incorporating our method in novel algorithms to reduce computational costs is another future research direction. 
\ack
The authors would like to thank Takashi Takahashi \color{black} and Masato Okada \color{black} for helpful insights. 
This work was partially funded by JSPS KAKENHI No. 17H00764 and JST CREST Grant Number JPMJCR1912, Japan (YK).
\appendix
\section{Derivation of equation \eref{tmp1}}
Herein, we derive  \eref{tmp1}. Dropping all indices and substituting \eref{eq:likelihood} and \eref{param_v_fn} into \eref{u_int}, 
we obtain
\begin{align}
m_{\mu \to i}(\vb{u}) &\propto \int \dd \vb{v} \exp\qty[ -\frac{\beta}{2}(y - \vb{u}^\ten \vb{v})^2 - \frac{\beta}{2} \vb{v}^\ten C\vb{v} + \beta \vb{D}^\ten \vb{v} ] \nonumber\\
& = \abs{ C + \vb{u}\vb{u}^\ten }^{-1/2} \exp \qty[ \frac{\beta}{2} (\vb{D} + y\vb{u})^\ten (C + \vb{u}\vb{u}^\ten )^{-1}(\vb{D} + y\vb{u}) ]. \label{A1}
\end{align}
The term in the exponential in \eref{A1} is manipulated using the Sherman--Morrison formula:
\begin{align}
      \frac{1}{2}(\vb{D} + y\vb{u})^\ten &(C + \vb{u}\vb{u}^\ten )^{-1}(\vb{D} + y\vb{u}) = \frac{1}{2}(\vb{D} + y\vb{u})^\ten \qty( C^{-1} - \frac{C^{-1}\vb{u}\vb{u}^\ten C^{-1}}{1 + \vb{u}^\ten C^{-1} \vb{u}} )(\vb{D} + y\vb{u}) \nonumber \\
& = \frac{y^2}{2} \frac{\vb{u}^\ten C^{-1}\vb{u}}{1 + \vb{u}^\ten C^{-1} \vb{u}} - \frac{1}{2} \frac{(\vb{u}^\ten C^{-1}\vb{D})^2}{1 + \vb{u}^\ten C^{-1}\vb{u}} + y\frac{\vb{u}^\ten C^{-1}\vb{D}}{1 + \vb{u}^\ten C^{-1}\vb{u}} + {\rm Const.}, \label{A2}
\end{align}
where $\rm{Const.}$ represents the terms that are independent of $\vb{u}$. 
Note that since $\vb{u}^\ten C^{-1}\vb{u}/(1 + \vb{u}^\ten C^{-1} \vb{u}) = 1 - 1/(1 + \vb{u}^\ten C^{-1} \vb{u})$ in the first term, 
\eref{A2} is given by 
\begin{equation}
      -\frac{1}{2}\frac{(y - \vb{u}^\ten C^{-1}\vb{D})^2}{1 + \vb{u}^\ten C^{-1}\vb{u}} + {\rm Const.} \label{A3}
\end{equation}
Using \eref{A3} and the matrix determinant lemma $\abs{C + \vb{u}\vb{u}^\ten} = \abs{C}(1 + \vb{u}^\ten C^{-1}\vb{u}) $ in \eref{A1}, we obtain \eref{tmp1}. 

\section{Replica computation and its relation to population dynamics}
Handling the objective function of \eref{eq:objective} as the Hamiltonian, the partition function is computed as:\begin{equation}
      Z_\beta(L, U^0, V^0) = \int \dd U \dd V \exp \qty[ -\frac{\beta}{2} \sum_{i,j}L_{ij} (\vb{u}_i^\ten \vb{v}_j - (\vb{u}^0_i)^\ten \vb{v}_j^0 )^2 - \frac{\beta \lambda}{2}\sum_{i = 1}^N \norm{\vb{u}_i}^2 - \frac{\beta \lambda}{2}\sum_{j = 1}^M \norm{\vb{v}_j}^2   ],
\end{equation}
where $L = \qty{L_{ij}} \in \qty{0,1}^{N\times M}$ is a sparse binary matrix, and $U^0 = \qty(\vb{u}^0_1, \cdots \vb{u}_N^0)^\ten,V^0=  \qty(\vb{v}^0_1, \cdots \vb{v}_M^0)^\ten$ is the ground truth (planted) 
matrix. In particular, we are interested in the free energy at zero temperature $\phi = \lim_{\beta \to \infty} \lim_{N, M\to \infty}  (N\beta)^{-1}\log Z_\beta$, 
as this enables us to assess the macroscopic properties of the maximum of the posterior distribution specified by 
\eref{eq:likelihood} and \eref{eq:prior}. 
Here, the limit of $N, M\to \infty$ is taken such that the ratio $\alpha \equiv M /N$ is kept constant. 
The free energy $\phi$ is a random variable depending 
on the configuration of random variables $L$ and $U^0, V^0$. 
The configurational average of $\phi$, often referred to as the \textit{quenched} average,
 can be calculated using the replica method, which is based on the following 
identity:
\begin{equation}\label{eq:replica_trick}
\av{\log Z_\beta}_{L, U^0, V^0} = \lim_{n\to +0} \frac{1}{n} \log \av{Z^n_\beta}_{L, U^0, V^0},      
\end{equation}
where $\left \langle \cdots \right \rangle_{L, U^0, V^0}$ stands for the configurational average with respect to $L, U^0$, and $V^0$. 
To avoid the difficulty of calculating $\av{Z^n_\beta}_{L, U^0, V^0}$ for $n\in \mathbb{R}$, 
the configurational average of $Z_\beta^n$ is first calculated 
as an analytic form of $n\in \mathbb{N}$, and the limit of $n\to +0$ is taken via analytical continuation to $n\in \mathbb{R}$. \\
\indent Herein, we derive the replica symmetric solution along the lines of \cite{Murayama00} and \cite{Kabashima12}. 
While we consider only the case where $L$ is a matrix with $c$ nonzero matrix elements per column and $r$ nonzero elements per row (so $Nr = Mc$),
 generalization to other binary masks is straightforward. 
 The planted matrices $U^0, V^0$ are assumed to follow the distribution 
$\priorU{U^0} = \prod_{i = 1}^N \priorU{\vb{u}^0_i}, \priorV{V^0} = \prod_{j = 1}^M \priorV{\vb{v}^0_j}$. 
First, the average over all configurations of $L$ is given by 
\begin{align}
      \av{(\cdots)}_L &= \mathcal{N}^{-1} \myTr_{L} \  (\cdots)\  \delta\qty(\sum_{i}L_{ij}, c)\ \delta\qty(\sum_{j}L_{ij}, r) \nonumber \\
      & = \mathcal{N}^{-1} \prod_{i = 1}^N \qty(\oint \frac{\dd z_i}{2\pi\sqrt{-1}} z_i^{c+1} )\prod_{j = 1}^M \qty(\oint \frac{\dd w_j}{2\pi\sqrt{-1}}w_j^{r+1})\ \myTr_{L}\  (\cdots)\ (z_iw_j)^{L_{ij}},
\end{align}
Where we used the integral representation of the Kronecker delta. Here, $\myTr_{L_{ij}} = \prod_{i,j} \sum_{L_{ij} = 0,1}$, and $\mathcal{N}$ is 
the total number of configurations, which can be calculated using the saddle point method as \cite{Murayama00} \[
 \mathcal{N} = \exp\qty(Nr\log Nr  -Nr - N\log r! - M\log c!).
 \]
Taking the $n-$replicated partition function indexed by $a = 1, \cdots , n$, and averaging over $L, U^0,$ and $V^0$ offers 
\begin{align}
      \label{eq:quenched}
&\av{Z^n_\beta}_{C, U^0, V^0} = \mathcal{N}^{-1} \prod_{i = 1}^N \qty( \oint \frac{\dd z_i}{2\pi \sqrt{-1}} z_i^{c + 1} ) \prod_{i = j}^M \qty( \oint \frac{\dd w_j}{2\pi \sqrt{-1}} w_j^{r + 1} ) \int \prod_{a = 0}^n \dd U^a \dd V^a \priorU{U^0} \priorV{V^0} \nonumber \\
&\quad \times \myTr_{L}\prod_{i,j} \qty{ (z_iw_j)^{L_{ij}} } \exp\qty[ -\frac{\beta}{2} \sum_{i,j}L_{ij} \sum_{a= 1}^n ( (\vb{u}^a_i)^\ten \vb{v}^a_j - (\vb{u}^0_i)^\ten \vb{v}_j^0 )^2 - \frac{\beta \lambda}{2} \sum_{a,i} \norm{\vb{u}^a_i}^2 - \frac{\beta \lambda}{2} \sum_{a,j} \norm{\vb{v}^a_j}^2 ].
\end{align}
The summation over $L_{ij}$ can be evaluated as 
\begin{align}
      \label{eq:approx_auxillary}
      &\quad \myTr_{L}\prod_{i,j} \qty{ (z_iw_j)^{L_{ij}} } \exp\qty[ -\frac{\beta}{2} \sum_{i,j}L_{ij} \sum_{a= 1}^n ( (\vb{u}^a_i)^\ten \vb{v}^a_j - (\vb{u}^0_i)^\ten \vb{v}_j^0 )^2] \nonumber \\
      & = \prod_{ij}\qty{ 1 + z_iw_j \exp\qty[ -\frac{\beta}{2} \sum_{a= 1}^n ( (\vb{u}^a_i)^\ten \vb{v}^a_j - (\vb{u}^0_i)^\ten \vb{v}_j^0 )^2] } \nonumber \\
      & = \exp \qty{ \sum_{i,j} \log \qty[ 1 + z_iw_j \exp\qty[ -\frac{\beta}{2} \sum_{a= 1}^n ( (\vb{u}^a_i)^\ten \vb{v}^a_j - (\vb{u}^0_i)^\ten \vb{v}_j^0 )^2] ]} \nonumber \\
      & \simeq \exp \qty{ \sum_{ij} z_iw_j \exp\qty[ -\frac{\beta}{2} \sum_{a= 1}^n ( (\vb{u}^a_i)^\ten \vb{v}^a_j - (\vb{u}^0_i)^\ten \vb{v}_j^0 )^2]}.
\end{align}
Now, define order parameter functions as 
\begin{align}
      \label{eq:order_param}
    Q( \qty{\vb{u}^a} ) = \frac{1}{N} \sum_{i = 1}^N z_i \prod_{a = 0}^n \delta( \vb{u}_i^a - \vb{u}^a ), \quad
    q( \qty{\vb{v}^a} ) = \frac{1}{M} \sum_{j = 1}^M w_j \prod_{a = 0}^n \delta( \vb{v}_j^a - \vb{v}^a ),
\end{align}
and their conjugate functions, denoted by $\hat{Q} , \hat{q}$, to constrain $Q, q$ to obey the above relations.
 More explicitly, we use the equality 
\begin{align}
      \label{eq:unity_Q}
    1 &= \int \qty[DQ] \ \delta\qty( \frac{1}{N} \sum_{i = 1}^N z_i \prod_{a = 0}^n \delta( \vb{u}_i^a - \vb{u}^a ) - Q( \qty{\vb{u}^a} ) )  \nonumber\\
 & = \int \qty[DQ]\qty[D\hat{Q}] \exp \qty[ -\int \dd \qty{\vb{u}^a}\  N Q( \qty{\vb{u}^a} ) \hat{Q}( \qty{\vb{u}^a} ) + \int \dd \qty{\vb{u}^a} \ \hat{Q}( \qty{\vb{u}^a} )  \sum_{i = 1}^N z_i \prod_{a = 0}^n  \delta( \vb{u}_i^a - \vb{u}^a )  ] ,
\end{align} 
and 
\begin{align}
      \label{eq:unity_q}
    1 &= \int \qty[Dq] \ \delta\qty( \frac{1}{M} \sum_{i = j}^M w_j \prod_{a = 0}^n \delta( \vb{v}_j^a - \vb{v}^a ) - q( \qty{\vb{v}^a} ) )  \nonumber\\
 & = \int \qty[Dq]\qty[D\hat{q}] \exp \qty[ -\int \dd \qty{\vb{v}^a}\  M q( \qty{\vb{v}^a} ) \hat{q}( \qty{\vb{v}^a} ) + \int \dd \qty{\vb{v}^a} \ \hat{q}( \qty{\vb{v}^a} )  \sum_{i = 1}^M w_j \prod_{a = 0}^n  \delta( \vb{v}_j^a - \vb{v}^a )  ] ,
\end{align} 
which is verified by employing results from functional integration. 
Using \eref{eq:order_param}, \eref{eq:approx_auxillary} is expressed as 
\begin{align}
      \label{eq:auxillary}
    &\sum_{i,j}z_iw_j \exp \qty{-\frac{\beta}{2} \sum_{a= 1}^n ( (\vb{u}^a_i)^\ten \vb{v}^a_j - (\vb{u}^0_i)^\ten \vb{v}_j^0 )^2  } \nonumber \\
    &= NM \int \dd \qty{\vb{u}^a}\dd \qty{\vb{v}^a}  \ Q(\qty{\vb{u}^a})q(\qty{\vb{v}^a}) \exp \qty[ -\frac{\beta}{2} \sum_{a = 1}^n ( (\vb{u}^a)^\ten \vb{v}^a - (\vb{u}^0)^\ten \vb{v}^0 )^2 ].
\end{align}
Inserting equality \eref{eq:unity_Q} and \eref{eq:unity_q}, and using \eref{eq:approx_auxillary} and \eref{eq:auxillary} to \eref{eq:quenched}, 
the integral over $z_i$ and $w_j$ can be performed. Employing the saddle point method with respect to the order parameter functions $Q, q, \hat{Q},$ and $\hat{q}$ 
offers the logarithm of the configurational average of $Z_\beta^n$:
\begin{align}
      \label{eq:log_partition}
    \frac{1}{N} \log &\av{Z^n_\beta}_{C, U^0, V^0} = \extr_{Q, q, \hat{Q}, \hat{q}}\left\{ M \int \dd \qty{\vb{u}^a} \dd \qty{\vb{v}^a} Q(\qty{\vb{u}^a}) q(\qty{\vb{v}^a}) \exp \qty[- \frac{\beta}{2}\sum_{a = 1}^n ( (\vb{u}^a)^\ten \vb{v}^a - (\vb{u}^0)^\ten \vb{v}^0 )^2 ] \right. \nonumber\\
    &\quad  \left. -\int \dd \qty{\vb{u}^a} Q(\qty{\vb{u}^a})\hat{Q}(\qty{\vb{u}^a})  - \alpha \int \dd \qty{\vb{v}^a} q(\qty{\vb{v}^a})\hat{q}(\qty{\vb{v}^a}) \right. \nonumber\\ 
    &\quad  \left. + \log \int \dd \qty{\vb{u}^a} \priorU{\vb{u}^0} \hat{Q}^c (\qty{\vb{u}^a}) \exp\qty( -\frac{\beta\lambda}{2} \sum_{a = 1}^n \norm{\vb{u}^a}^2 ) \right.\nonumber \\
    &\quad \left.+  \alpha \log \int \dd \qty{\vb{v}^a} \priorV{\vb{v}^0} \hat{q}^r (\qty{\vb{v}^a}) \exp\qty( -\frac{\beta\lambda}{2} \sum_{a = 1}^n \norm{\vb{v}^a}^2 ) - r\log(Nr) + r  \right\}.
\end{align}
Now, we assume that the saddle point is dominated by the order parameter functions of the form 
\begin{align}
    Q(\qty{\vb{u}^a}) &= P_U(\vb{u}^0) \int \qty[Df] \ \rho_U(f|\vb{u}^0) \prod_{a = 1}^n \frac{e^{ -\beta f(\vb{u}^a)}}{\int \dd \vb{u} e^{-\beta f(\vb{u})} }, \\
    q(\qty{\vb{v}^a}) &= P_V(\vb{v}^0) \int \qty[Dg] \ \rho_V(g|\vb{v}^0) \prod_{a = 1}^n \frac{e^{ -\beta g(\vb{v}^a)}}{\int \dd \vb{v} e^{-\beta g(\vb{v})} }, \\
    \hat{Q}(\qty{\vb{u}^a}) &= \hat{P}_U(\vb{u}^0) \int  [D\hat{f}]\  \hat{\rho}_U(\hat{f}|\vb{u}^0)\exp\qty( -\beta \sum_{a = 1}^n \hat{f}(\vb{u}^a) ),\\
    \hat{q}(\qty{\vb{v}^a}) &= \hat{P}_V(\vb{v}^0) \int \qty[D\hat{g}]\ \hat{\rho}_V(\hat{g}|\vb{v}^0)\exp\qty( -\beta \sum_{a = 1}^n \hat{g}(\vb{v}^a) ).
\end{align}
Here, functionals $\rho_U, \rho_V, \hat{\rho}_U,$ and $\hat{\rho}_V$ are distributions of functions, and $P_U, P_V, \hat{P}_U,$ and $\hat{P}_V$ are 
potentially non-normalized functions. 
This form of order parameter functions is induced from the replica symmetric ansatz \cite{Nishimori01}, where $Q, q, \hat{Q}$, and $\hat{q}$ are 
functions symmetric with respect to permutations of replica indices $a = 1, \cdots, n$. 
Inserting these expressions to \eref{eq:log_partition}, and extremizing with respect to functions $P_U, P_V, \hat{P}_U,$ and $\hat{P}_V$ 
in the limit $n\to 0$ offers
\begin{equation}
      \label{eq:normalization}
\hat{P}_U\hat{P}_V = Nc, \quad P_U(\vb{u}^0)\hat{P}_U = c\priorU{\vb{u}^0}, \quad P_V(\vb{v}^0)\hat{P}_V = r\priorV{\vb{v}^0}.    
\end{equation}
Using \eref{eq:normalization} and the Laplace method to evaluate the integrals in the limit $\beta \to \infty$ for \eref{eq:log_partition}, we obtain the 
free energy as  
\begin{align}
    &\frac{\phi}{c} = \lim_{\beta \to \infty} \frac{1}{\beta} \lim_{N\to \infty}  \lim_{n\rightarrow +0}\frac{1}{nNc} \log \av{Z^n(\beta, C)}_{C, U^0, V^0} \nonumber \\ 
    & = \extr_{\rho_U, \rho_V, \hat{\rho}_U, \hat{\rho}_V}\left\{ -\av{ \min_{\vb{u}, \vb{v}} \qty[\frac{1}{2} (\vb{u}^\ten \vb{v} + (\vb{u}^0)^\ten \vb{v}^0)^2 + f(\vb{u}) + g(\vb{v}) ] - \min_{\vb{u}}f(\vb{u}) - \min_{\vb{v}}g(\vb{v}) }_{\vb{u}^0, \vb{v}^0, f,g } \right.\nonumber \\
    &\quad  \left. + \av{ \min_{\vb{u}}(f(\vb{u}) + \hat{f}(\vb{u})) - \min_{\vb{u}}f(\vb{u}) }_{\vb{u}^0,f, \hat{f}} + \av{ \min_{\vb{v}}(g(\vb{v}) + \hat{g}(\vb{v})) - \min_{\vb{v}}g(\vb{v}) }_{\vb{v}^0, g, \hat{g}}  \right. \nonumber\\ 
    &\quad  \left. - \frac{1}{c} \av{ \min_{\vb{u}} \qty[ \frac{\lambda}{2}\norm{\vb{u}}^2 + \sum_{k = 1}^c \hat{f}_k(\vb{u}) ] }_{ \vb{u}^0, \qty{\hat{f}_k}_{k = 1}^c } - \frac{1}{r} \av{ \min_{\vb{v}} \qty[ \frac{\lambda}{2}\norm{\vb{v}}^2 + \sum_{l = 1}^r \hat{g}_k(\vb{v}) ] }_{ \vb{v}^0, \qty{\hat{g}_l}_{l = 1}^r } \right\}.
\end{align}
where $\left \langle \cdots \right \rangle_{u^0, v^0, f, g} $represents the average operation with respect to $\priorU{\vb{u}^0}$, $\priorV{\vb{v}^0}$, $\rho(f|\vb{u}^0)$, $\rho(g|\vb{v}^0)$, and similarly for the other brackets. 
By taking the function derivative with respect to $\rho_U, \rho_V, \hat{\rho}_U, \hat{\rho}_V$, it can be confirmed that 
the equations of state which describe the replica symmetric solution are given by the following:
\begin{align}
    \hat{\rho}_U(\hat{f}|\vb{u}^0) &= \av{ \delta \qty(\hat{f}(\vb{u}) - \min_{\vb{v}} \qty[ \frac{1}{2}(\vb{u}^\ten \vb{v} - (\vb{u}^0)^\ten \vb{v}^0)^2+g(\vb{v}) ] +\min_{\vb{v}}g(\vb{v}))}_{\vb{v}^0, g}, \label{eq:rho_f}\\
    \hat{\rho}_V(\hat{g}|\vb{v}^0) &= \av{ \delta \qty(\hat{g}(\vb{v}) - \min_{\vb{u}} \qty[ \frac{1}{2}(\vb{u}^\ten \vb{v} - (\vb{u}^0)^\ten \vb{v}^0)^2+f(\vb{u}) ] +\min_{\vb{u}}f(\vb{u}))}_{\vb{u}^0, f},\label{eq:rho_g}\\
    \rho_U(f|\vb{u}^0) &= \av{\delta \qty(f(\vb{u}) - \sum_{k = 1}^{c-1}\hat{f}_k(\vb{u}) - \frac{\lambda}{2} \norm{\vb{u}}^2 )}_{\vb{v}^0, \qty{\hat{f}_k}_{k = 1}^{c-1}},\label{eq:rho_fh}\\
    \rho_V(g|\vb{v}^0) &= \av{\delta \qty(g(\vb{v}) - \sum_{l = 1}^{r-1}\hat{g}_l(\vb{v}) - \frac{\lambda}{2} \norm{\vb{v}}^2 )}_{\vb{u}^0, \qty{\hat{g}_l}_{l = 1}^{r-1}}.\label{eq:rho_gh}
\end{align}
These can easily be augmented to the noisy case by adding a noise term to $(\vb{u}^0)^\ten \vb{v}^0$ and averaging over its distribution in 
\eref{eq:rho_f} and \eref{eq:rho_g}.\\
\indent The system of equations \eref{eq:rho_f}--\eref{eq:rho_gh} 
determine the probability density of functions $f, g, \hat{f}$ and $\hat{g}$, 
which correspond to cavity biases and cavity distributions in the BP algorithm. 
Thus, 
the replica symmetric solution describes, by nature, 
the typical properties of cavity biases and cavity distributions in the form of 
its distribution on functional space. 
PD aims to solve this set of equations by preparing a population of $f, g, \vb{u}^0$ and $\vb{v}^0$, 
and performing Monte Carlo sampling for \eref{eq:rho_f}--\eref{eq:rho_gh}. 
The empirical distribution of $f$ and $g$ 
obtained from the population after sufficient sampling 
iterations approximates the fixed point of the equations of state 
in the large population size limit \cite{MM09}. 
Again, the problem of continuous degrees of freedom persists; this is avoided by 
approximating the functions by a Gaussian, which reduces to the PD algorithm in \ref{subsection:PD}. 
\section*{References}
 \bibliographystyle{iopart-num}
 \bibliography{ref}
 \end{document}